\documentclass[dvipsnames]{article} 
\usepackage{iclr2025_conference,times}

\usepackage{amsmath,amsfonts,bm}

\def\eqref#1{equation~\ref{#1}}

\def\1{\bm{1}}

\DeclareMathAlphabet{\mathsfit}{\encodingdefault}{\sfdefault}{m}{sl}
\SetMathAlphabet{\mathsfit}{bold}{\encodingdefault}{\sfdefault}{bx}{n}

\usepackage{hyperref}
\usepackage{url}
\usepackage{enumitem}
\usepackage{amsmath}
\usepackage{graphicx}
\usepackage{relsize}
\usepackage{pifont}
\usepackage{multirow}
\usepackage{amssymb}
\usepackage{amsthm}
\usepackage{multirow}       
\usepackage{colortbl}       
\usepackage{graphicx}       
\usepackage{booktabs, tabularx, xcolor, colortbl}
\usepackage{siunitx}
\usepackage{caption}
\usepackage{adjustbox}
\newcolumntype{Y}{>{\centering\arraybackslash}X}

\title{Open Data Synthesis For Deep Research
}

\iclrfinalcopy

\author{%
  Ziyi Xia\textsuperscript{\thanks{Equal contribution.}} \quad
  Kun Luo\textsuperscript{\footnotemark[1]} \quad
  Hongjin Qian\textsuperscript{\footnotemark[1]} \quad
  Zheng Liu\textsuperscript{\thanks{Corresponding Author}} \\
  BAAI \\
  \texttt{\{ziyixia85,luokun695,chienqhj,zhengliu1026\}@gmail.com}\\
}

\begin{document}

\maketitle

\begin{abstract}
Large language models (LLMs) are increasingly expected to go beyond simple factual queries toward Deep Research—tasks that require decomposing questions into sub-problems, coordinating multi-step reasoning, and synthesizing evidence from diverse sources. We formalize Deep Research tasks with verifiable answers as Hierarchical Constraint Satisfaction Problems (HCSPs), which are fundamentally different from single-constraint, multi-hop, or flat CSP formulations. However, existing benchmarks (e.g., Natural Questions, HotpotQA) fail to capture this complexity, while recent synthetic datasets often introduce shortcut reasoning, knowledge leakage, or lack sufficient structural depth.

To address this gap, we introduce InfoSeek, a scalable framework for synthesizing complex Deep Research tasks. InfoSeek uses a dual-agent system to recursively build a Research Tree from large-scale webpages, blurring intermediate nodes into valid sub-problems, and converting these trees into natural language questions that require traversing the full hierarchy.  It also enables rapid scaling, yielding over 50K training examples, a curated test set, and reasoning trajectories generated via reject sampling.
Experiments show that models trained on InfoSeek consistently outperform strong baselines. On a challenging benchmark BrowseComp-Plus, 3B LLMs optimized with InfoSeek surpass much larger 32B models and lightweight commercial APIs (e.g., Gemini2.5-Flash), while achieving performance comparable to stronger APIs (e.g., Gemini2.5-Pro). By preserving meta-information such as intermediate steps and retrieval labels, InfoSeek further supports advanced optimization strategies, including compound reward design and trajectory-level exploration. We provide our codes and datasets in \href{https://github.com/VectorSpaceLab/InfoSeek}{this repository}.
\end{abstract}

\begin{figure}[h]
    \centering
    \includegraphics[width=0.82\linewidth]{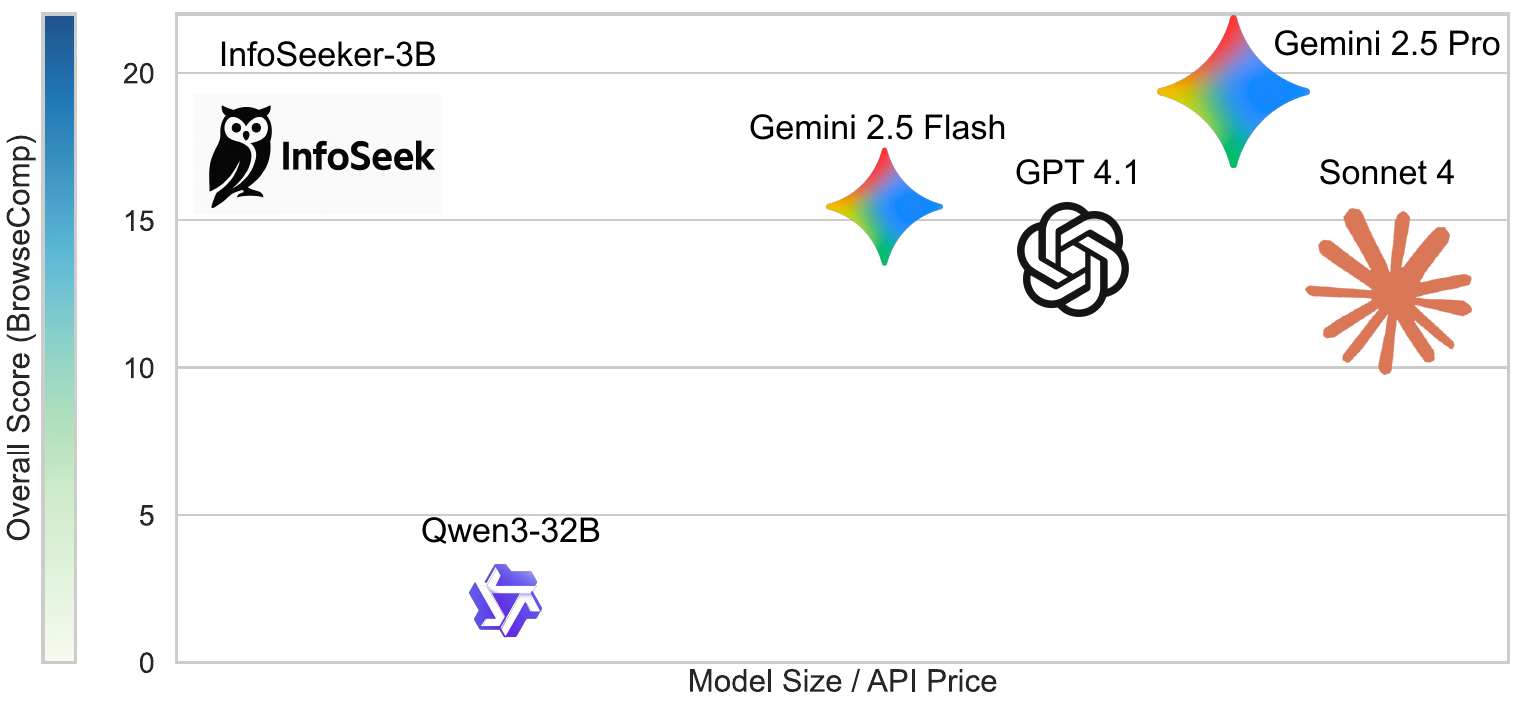}
    \caption{Performance comparison on the BrowseComp-Plus benchmark. InfoSeeker-3B, a compact LLM trained with the InfoSeek dataset, significantly outperforms Qwen3-32B and achieves performance on par with leading commercial LLMs (ordered in API prices), highlighting the strong potential of InfoSeek for advancing Deep Research tasks.
Notably, the InfoSeek data synthesis framework is fully open-source, enabling convenient and scalable dataset construction.}
    \label{fig:compare}
\end{figure}

\section{Introduction}

Recently, large language models (LLMs) have transformed AI by generating and interpreting language with unprecedented fluency and contextual depth~\citep{gpt-4,gemini25}. Beyond simple factual queries, emerging advances point to a new frontier: \emph{Deep Research}—where models must decompose complex tasks, generate sub-queries, and reason across diverse sources of information~\citep{openai2025deepresearch}. Unlike conventional information seeking, which suffices for straightforward retrieval, Deep Research demands synthesizing heterogeneous evidence, coordinating multi-step reasoning, and often interacting with external tools~\citep{wu2025agentic,zhang2024agentic}. Such capabilities are essential for domains like scientific discovery and policy analysis, where problems are open-ended and knowledge landscapes continuously evolve. Consequently, Deep Research is increasingly viewed as a cornerstone for the next generation of LLMs, shifting them from conversational assistants to autonomous knowledge engines~\citep{li2025towards}.

A \emph{deep research question} goes beyond simple factual lookup~\citep{openai2025deepresearch}. It requires navigating multiple layers of knowledge integration, and is best understood in contrast with several simpler problem types.  
A constraint satisfaction problem is solved by combining several independent conditions to narrow the candidate set. A multi-hop problem demands a sequence of dependent inferences and search~\citep{HotpotQA,zhao2024retrievalaugmentedgenerationrag,qian2025hawkbench}.  
Deep research questions extend beyond both by involving a hierarchy of interdependent constraints that intertwine both parallel conditions and sequential steps. The solution emerges only through progressively resolving this hierarchy of sub-questions. When the final answer is unique and verifiable, the reasoning process can be naturally represented as a tree, with intermediate vertices denoting sub-questions and branches encoding their logical dependencies.  
In this work, rather than long-form open-ended tasks such as report writing, we focus on deep research questions that yield a unique and verifiable answer.

\begin{table}[t]
\centering
\caption{Comparison of the open-source status of classical QA datasets and recent data synthesis approaches for Deep Research. While prior datasets either lack structural depth or remain limited in scale, InfoSeek provides the first large-scale dataset dedicated to Deep Research scenarios, capable of generating hierarchical constraint satisfaction problems with controllable complexity, and supporting easy scalability for diverse research needs.}
\label{tab:dataset-comparison}
\renewcommand{\arraystretch}{1.2}
{\setlength{\tabcolsep}{6pt}%
\begin{tabularx}{0.95\linewidth}{@{} >{\raggedright\arraybackslash}X
                                    c
                                    c 
                                    c
                                    c
                                    c @{}}
\toprule
\textbf{Name} & \textbf{Problem} & \textbf{Data Source} & \textbf{QA pairs} & \textbf{Trajectories} & \textbf{Framework} \\
\midrule
NQ                 & Single-hop & Wiki & {300k+} & {--}     & {--} \\
HotpotQA           & Multi-hop  & Wiki & {100k+} & {--}     & {--} \\
\midrule
WebWalkerQA        & Multi-hop  & Web & 14.3k   & {--}     & {--} \\
InForage           & Multi-hop  & Web & {--}    & {--}     & {--} \\
SimpleDeepSearcher & Multi-hop  & {--} & {--}    & 871     & Open \\
Pangu DeepDiver    & Multi-hop  & Web & {--}    & {--}     & {--} \\
WebDancer          & Multi-hop  & Wiki\&Web & 200     & 200     & {--} \\
WebShaper          & Complex    & Wiki & 500     & {--}     & {--} \\
\rowcolor{gray!12} InfoSeek     & HCSP  & Wiki\&Web      & {50k+}    & 16
.5k   & Open \\
\bottomrule
\end{tabularx}}
\end{table}

Some recent approaches propose carefully designed workflows for planning and tool use~\citep{searcho1,wu2025agentic,soni2025coding,zhang2025agentorchestra}. While effective in narrow domains, these workflows lack the flexibility required for diverse Deep Research tasks~\citep{li2025towards}.
Another line of work enhances models’ reasoning and search capabilities through supervised fine-tuning~\citep{sun2025simpledeepsearcher} or reinforcement learning~\citep{searchr1,r1searcher,zheng2025deepresearcher}.
Although such methods achieve gains on traditional and multi-hop QA benchmarks, their training still depends heavily on datasets like Natural Questions~\citep{kwiatkowski2019natural} and HotpotQA~\citep{HotpotQA}, which remain far simpler than real Deep Research scenarios.
Other efforts~\citep{wu2025webwalker,shi2025pangu} attempt to construct open-source QA datasets or trajectories, but these remain focused on multi-hop QA. 
More recent studies~\citep{qian2025scent,shi2025pangu,wu2025webdancerautonomousinformationseeking,tao2025webshaper} explore harder question types involving web pages or Wikipedia, yet neither their datasets nor their workflows are publicly released. Table~\ref{tab:dataset-comparison} compares the open-source availability of classical QA datasets and recent data synthesis efforts for Deep Research. The results highlight a scarcity of high-quality, large-scale datasets explicitly designed for Deep Research in the open-source community.

\begin{figure}
    \centering
    \includegraphics[width=\linewidth]{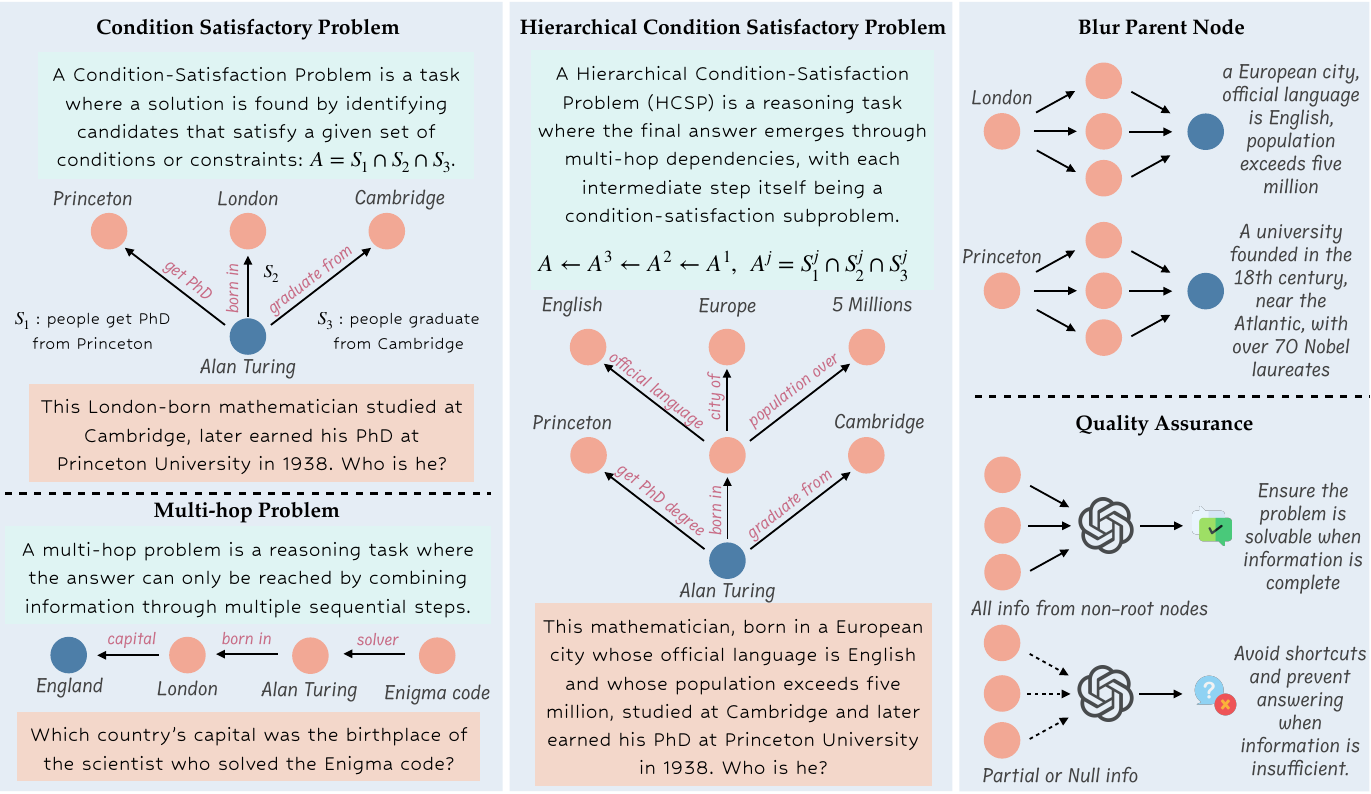}
    \caption{Illustration of Constraint Satisfaction Problems (CSP), Multi-hop Problems (MHP), and Hierarchical Constraint Satisfaction Problems (HCSP), the latter formalized as Deep Research questions with unique, verifiable answers. The right panel depicts the blurred parent node technique and quality assurance process employed in our InfoSeek data synthesis framework.}
    \label{fig:idea}
\end{figure}

To overcome the limitations of existing benchmarks, we propose \textbf{InfoSeek}, a novel data synthesis paradigm that generates structurally complex Deep Research questions and answers from large-scale unstructured text. As shown in Fig~\ref{fig:idea}, the process begins by mining entities and their relations from raw text corpora. Leveraging these relations, we randomly sample a root entity and recursively construct a \emph{Research Tree}, where the root denotes the final answer, internal vertices represent intermediate sub-problems, and edges encode logical dependencies among entities.  
Accompany with adding entities, we enrich the tree structure by blurring the parent vertices with more constraints. This helps enhance the difficulty and ensure the uniqueness of the answer to the corresponding question. Through the process, every internal vertex naturally becomes a constraint satisfaction problem (CSP), and only by solving all sub-questions and integrating multiple such layers can the reasoning path eventually converge to the root answer.  
To convert these structured trees into natural language questions, we provide the blurred vertex descriptions to a powerful LLM such as DeepSeek V3 or GPT-4.1. The model is prompted to generate queries whose solution requires traversing the entire tree. This ensures that each synthesized question enforces genuine multi-step reasoning, prevents potential shortcut and yields a unique, verifiable answer grounded in factual evidence.  
By design, InfoSeek produces datasets that are structurally diverse, complexity-controllable, and intrinsically verifiable, providing a scalable foundation for training and evaluating Deep Research agents.

To assess whether InfoSeek supports model optimization, we conduct supervised fine-tuning and reinforcement learning experiments. An SFT trajectories dataset is first constructed via rejection sampling to ensure correctness, followed by standard RL training on InfoSeek to balance exploration and exploitation. Under this pipeline, models trained with InfoSeek already outperform strong baselines. In addition, since InfoSeek preserves meta-information such as intermediate steps and retrieval labels, it offers valuable signals for designing more sophisticated RL rewards, which we leave for future work.

In brief, the work present three contributions: 
(1) We formalize Deep Research questions with verifiable answers as \emph{Hierarchical Constraint Satisfaction Problems} (HCSPs), and provide a principled distinction from simpler multi-hop and flat CSP problem types.  
(2) We introduce \emph{InfoSeek}, an autonomous and scalable data synthesis framework that adheres to this definition, enabling high-quality dataset construction with explicit control over structural complexity and principled scalability.  
(3) Following this framework, we build a large-scale Deep Research dataset that fully records the construction process. Empirically, we verify the effectiveness of InfoSeek by fine-tuning and optimizing models on the dataset, achieving performance that consistently surpasses strong baselines.  
(4) The dataset with more than 50k QA pairs, 16.5k reasoning trajectories, and data construction framework of InfoSeek are fully open-sourced. Thus encouraging further research or development in the community.

\section{Preliminary}
\label{sec:prelim}

In this section, we lay out the foundational concepts required to formalize deep research questions. We begin by classifying fundamental problem types that serve as building blocks for more complex reasoning tasks. These classifications provide the basis for defining Hierarchical Constraint Satisfaction Problems (HCSPs), a framework that captures the layered structures and interdependent constraints characteristic of deep research. Finally, we present a method for constructing HCSPs systematically from tree representations.

\subsection{Fundamental Problem Types}
\textbf{Constraint Satisfaction Problem (CSP).} 
A group of simple research questions can be formalized as a constraint satisfaction problem, where the goal is to identify the unique answer set \(A\) that simultaneously satisfies all constraints extracted from the question. Formally, given a set of constraints \(C_q = \{c_1, c_2, \dots, c_n\}\), we define
\begin{equation}
    A = \bigcap_{i=1}^n S(c_i) \quad \text{s.t. } |A| = 1, \; |S(c_i)| \geq 1 \;\; \forall i ,
\end{equation}
where \(S(c_i)\) denotes the set of entities that satisfy constraint \(c_i\).  
When \(n=1\), the CSP reduces to the base case of a \emph{single-constraint problem}, which involves exactly one condition and yields a unique ground-truth answer. For example, the question ``Who developed the theory of relativity?'' corresponds to \(C_q=\{c_1\}\) with \(c_1=\) ``developed the theory of relativity'', whose solution is \(A=\{\text{Albert Einstein}\}\).  
For \(n>1\), such as in Fig.~\ref{fig:idea}.a with constraints \(c_1:\) got PhD from Princeton University in 1938, \(c_2:\) born in London, and \(c_3:\) graduated from University of Cambridge. The intersection $S(c_1)\cap S(c_2)\cap S(c_3)$ yields the correct answer \(A=\{\text{Alan Turing}\}\). Solving a CSP thus requires both accurate retrieval of entities for each constraint and reasoning over their intersection.

\textbf{Multi-hop Problem (MHP).}  
A multi-hop problem is a reasoning task where the answer can only be obtained by sequentially chaining together multiple inference steps, with each step depending on the output of the previous one. Formally, given an initial constraint \(c\), the solution is derived through a composition of reasoning functions:
\begin{equation}
    A = S^{(k)}(c) = \underbrace{S \circ S \circ \dots \circ S}_{k \;\text{times}} (c),
\end{equation}
where \(k\) denotes the number of reasoning hops.  

Take the 3-hop question illustrated in Fig.~\ref{fig:idea}.b as an example:  
(1) starting with \(c=\) ``scientist who solved the Enigma code'', we obtain \(S(c)=\{\text{Alan Turing}\}\);  
(2) using this entity, we resolve \(S(\text{``birthplace of Alan Turing''})=\{\text{London}\}\);  
(3) finally, the problem reduces to a single-constraint query: ``which country has London as its capital'', yielding \(A=\{\text{England}\}\).  

Unlike constraint satisfaction problems, where the answer arises from intersecting parallel constraints, multi-hop problems require strictly ordered reasoning. A central challenge is that errors at intermediate steps propagate forward, potentially invalidating the final result.

\subsection{Definition of Deep Research Task}
A \emph{deep research task}~\citep{google2024deepresearch, openai2025deepresearch, perplexity2025deepresearch} is a complex information-seeking activity characterized by multi-layered information dependencies. It evaluates the ability of an agent—whether a single LLM or a system of collaborating LLMs—across several dimensions: (1) multi-step reasoning, (2) decomposition of a complex question into manageable sub-questions, (3) strategic query generation and iterative search for relevant information, and (4) integration of evidence from multiple sources into a coherent final output.  
In this work, we study complex question answering as a representative instance of deep research, since it naturally supports both verifiability and reproducibility.  

Building on fundamental problem types, we formalize a deep research question as a \textbf{Hierarchical Constraint Satisfaction Problem (HCSP)}. In an HCSP, the final answer is not directly accessible but must be progressively uncovered by satisfying a hierarchy of interdependent constraints. Solving such problems requires systematically pruning the search space at each level—eliminating candidates inconsistent with the accumulated evidence—until the root level converges to a unique valid answer.  
Formally, given a question \(x\) with a set of constraints \(C_x=\{c_1,\dots,c_k\}\) and a set of sub-questions \(Y_x=\{y_1,\dots,y_m\}\), we define a hierarchical decomposition \(H(\cdot)\) as:
\begin{equation}
    H(x) = \bigcap_{i=1}^{k} S(c_i) \;\cap\; \bigcap_{j=1}^{m} H(y_j), 
    \quad \text{with } \bigcap \varnothing := \mathbb{U}, \label{eq:4}
\end{equation}
where \(\mathbb{U}\) denotes the universal set. 
The final answer \(A\) of a hierarchical constraint satisfaction problem $q_H$ is then given by
$
    A = H(q_H).
$

In this framework, both constraint satisfaction problem and multi-hop problem emerge as special cases of HCSP. Unlike classical CSPs that operate on flat, independent constraints, HCSPs impose structured, multi-level reasoning: the validity of higher-level conclusions depends on satisfying all lower-level constraints. This hierarchical pruning process not only parallels algorithmic paradigms such as constraint propagation in AI, but also echoes human reasoning, where complex judgments arise from integrating multiple, interdependent strands of evidence.

\subsection{Constructing HCSP from Research Tree}

With a clear definition of HCSP in hand, we now argue that every HCSP admits an underlying tree structure. In this subsection, we first formalize the research tree representation and the basic operations for constructing it. We then explain how to derive an HCSP from a given research tree, and finally discuss potential issues that may arise during this construction.

\subsubsection{A Research Tree from Entities}

In graph theory~\citep{bondy1976graph}, a \emph{tree} is defined as a connected, acyclic graph. Or equivalently, a tree
$
    T=(V,E)
$
is a graph in which, for any two vertices $u,v \in V$, there exists exactly one simple path between them.  
Here we define a \emph{research tree} $\mathcal{T}=(V,E)$, where each vertex $v \in V$ represents a knowledge entity (e.g., ``Alan Turing'', ``University of Cambridge'') or trivial fact (e.g., ``1910s", ``summer of 1925"), and each edge $(v,w)\in E$ connects two vertices and represents their relationship (e.g., ``Alan Turing graduated from the University of Cambridge'', ``Alan Turing was born in 1910s).

A research tree can be constructed recursively as follows:
\begin{itemize}[leftmargin=*]
    \item
        \textbf{Base case:} A single vertex $r$ with no edges as the root:
        \begin{equation}
            \mathcal{T}=(\{r\}, \varnothing).
            \label{eq:5}
        \end{equation}
    \item
        \textbf{Recursive expansion:} Given a tree $\mathcal{T}=(V,E)$, we may expand it by introducing a new vertex $w\notin V$ and connecting it with exactly one edge to some existing vertex $v\in V$:
        \begin{equation}
            \mathcal{T}=(V\cup\{w\}, E\cup\{(v,w)\}).
            \label{eq:6}
        \end{equation}
\end{itemize}

\subsubsection{From Research Tree to HCSP}

Once a research tree has been defined, we can recursively construct an HCSP that reflects its hierarchical structure:

\begin{itemize}[leftmargin=*]
    \item
        \textbf{Base case:} For a node $v$ of height $1$ (i.e., all children $child(v)=\{w_1,\dots,w_n\}$ are leaves), each edge $(v,w_i)$ is converted into a constraint $c_i$. A question $q_v$ is then formed by:
        \begin{equation}
            q_v = Q(C_v), \quad \text{where } v = H(q_v), \; C_v=\{c_1,\dots,c_n\},
            \label{eq:7}
        \end{equation}
        with $Q(\cdot)$ denoting the function that combines constraints into a question, and $H(\cdot)$ the decomposition operator defined in Eq.~\ref{eq:4}.  
        This case reduces to constructing a standard CSP.
    \item
        \textbf{Recursive step:} For a node $v$ of height $\geq 1$, partition its children into leaves $\{w_1,\dots,w_k\}$ and internal nodes $\{w_{k+1},\dots,w_n\}$. Each edge $(v,w_i)$ with $i \leq k$ is converted into a constraint $c_i$, while each internal child $w_j$ with $j>k$ recursively yields a sub-question $Q(w_j)$. The resulting question can be formalized by:
        \begin{equation}
            q_v = Q\!\left(C_v \cup \{Q(w_j) \mid j=k+1,\dots,n\}\right).
            \label{eq:8}
        \end{equation}
\end{itemize}

Finally, for a research tree $\mathcal{T}$ with root $r$, the corresponding HCSP is obtained as:
$
    q = Q(r).
$

\subsection{Potential Issues in Tree-based HCSP Construction}

While the tree-based construction of HCSP provides a systematic framework, it also introduces two potential issues.  
\emph{First}, the problem may be \textbf{underdetermined}: even after combining multiple constraints, the answer set may remain non-unique, leaving ambiguity in the solution space.  
\emph{Second}, the problem may be \textbf{overdetermined}: in some cases, a single constraint (or a small subset of constraints) already yields a unique solution, leading to premature convergence and diminishing the role of hierarchical reasoning.  
Both phenomena challenge the intended multi-level structure of HCSP, making it challenge to synthesize high quality data.  
In practice, however, we adopt specific data construction techniques to mitigate these issues and preserve the richness of hierarchical reasoning.

%

\section{InfoSeek: Scalable Data Synthesis for Deep Research}






\label{sec:dataset}

In this section, we present \textbf{InfoSeek}, a novel data synthesis framework for Deep Research tasks. Building on the theoretical foundations in Sec.~\ref{sec:prelim}, we design a dual-agent framework that generates structurally complex Hierarchical Constraint Satisfaction Problems (HCSPs) in an automatic and scalable manner, and incorporates mechanisms to ensure the quality and controllability of the generated questions. We first discuss the designed actions for the two agents. Then we introduce how we ensure data quality. Lastly, we present the statistics of our constructed dataset.

The tree construction process is orchestrated by two cooperating agents incrementally based on webpages and the Wikipedia dump. The Planner maintains a global view of the partially constructed tree, selecting target vertices and specific actions according to global complexity objectives. This ensures balanced growth across sequential and parallel reasoning demands. The Browser executes the Planner’s chosen action by browsing webpages of the selected vertices, extracting candidate hyperlinks (for depth) or atomic claims (for constraints), and validating them for relevance. A complete iteration begins with the first step of action 1, follows by a series of actions alternating between action 2 and 3, and ends with action 4. Each step expansion is recorded with an explicit evidence trace, guaranteeing verifiability.

\subsection{Action 1: Initialization from Research Anchors}

The first step of InfoSeek data sythesis pipeline is to select an entity to create the first vertex as the root of the research tree. We use webpages and full Wikipedia dump as the underlying knowledge base. After filtering out invalid or trivial pages, each remaining page—identified by its title and containing sufficient content—is treated as an entity. Within each page, all mentioned entities (validated by the presence of a hyperlink) or facts are related vertices to the current entity vertex. And sentences that describe a relationship between the current entity and any other entities or facts are treated as candidate edges connecting them. 

At the beginning of each iteration, the Planner chooses action 1 to initiate the construction of the research tree. Specifically, the Browser sample a valid entity from the Wikipedia knowledge base to serve as the final answer, and creates a vertex $r$ as the root to form the very first research tree $\mathcal{T}_0$. The browser then performs the first extension by selecting a related entity to create a child vertex $w$ of $r$, along with the connecting edge $(r, w)$.

\subsection{Action 2: Blurring Parent with Constraints}
Given a research tree $\mathcal{T}_t$ at step $t$, action 2: blurring parent with constraints, is available to expand a specific vertex with constraints. The Planner agent first identifies a vertex $v$ whose current constraints are insufficient to uniquely derive $v$.
The Browser then selects $k$ claims from $v$'s webpage that, taken together, yield the unique and determinate answer $v$. These claims form the child vertices $w_1,\dots,w_k$, along with their corresponding edges.
To avoid the \textbf{overdetermination} issue, we ensure that the resulting candidate sets are mutually exclusive, i.e., without inclusion relations.

\subsection{Action 3: Extending the Tree}
To increase the depth of the research tree, the Planner can perform a vertical expansion by choosing action 3. Starting from an existing vertex $v$ that represents an entity, the Browser agent extracts a hyperlink from the corresponding article that indicates a dependency (e.g., “$v$ was discovered by $w$”). This operation creates a new child node $w$ of $v$, thereby extending the logical dependency chain. Such an expansion increases the height of the tree and yields a structure that requires the model to perform an additional step of reasoning.

\subsection{Action 4: Termination and Generation of the Question}
Unlike prior work that often relies on surface-level heuristics for difficulty, our framework directly controls and monitors the structural complexity of the research tree. Termination is triggered only when the research tree achieves the desired complexity and all vertices have sufficient constraints. At this point, the Planner agent constructs the final complete question based on the research tree. Because the expansion operations rely on lightweight rules (i.e., hyperlink and fact extraction), the process is highly scalable and enables the cost-efficient synthesis of large-scale datasets.

\subsection{Data Quality Assurance}
\label{subsec:quality}

To ensure our dataset is robust and effective for training, we instituted a rigorous, two-pronged quality assurance protocol focused on \textbf{Difficulty} and \textbf{Verifiability}.

The \textbf{Difficulty} criterion ensures that problems are non-trivial and cannot be solved by relying on a LLM's parametric memory alone. To validate this, we challenged Qwen2.5-32B-Inst~\citep{team2024qwen2} to answer the questions directly. Our validation confirms the dataset's high degree of challenge: the model was able to correctly answer only 2\% of the questions. We remove these samples to further enhance difficulty.

The \textbf{Verifiability} criterion confirms that each question is factually grounded and solvable via the generated search trajectory. We present Gemini 2.5 Flash~\citep{comanici2025gemini} API with the ground-truth web pages from the constructed search path, intermixed with a set of distractor documents. The LLM is then tasked with deriving the correct answer from this provided context. We filter out the questions with wrong answer, multiple possible answers, or unable to solve. This process filter out the questions that contain ambiguity or even unsolvable, effectively preventing the \textbf{underdetermined} issue.

\subsection{Statistics}
\label{subsec:quality}
As shown in Table~\ref{tab:vertex_analysis}, InfoSeek comprises more than 50K samples, with the total data curation cost as \$571.8, provided for reproducibility. The distribution concentrated on problems requiring 4 to 6 reasoning vertices. To further quantify the dataset's complexity, we measure the failure rate of a powerful baseline, Qwen2.5-72B~\citep{qwen2}, using CoT prompting~\citep{wei2022chain}. This approach serves as a reliable proxy for deep research difficulty, as established by prior work~\citep{browsecomp}.
The results reveal a high overall failure rate of 92.7\%, confirming that our dataset poses a significant challenge even for powerful models. Crucially, the failure rate exhibits a strong positive correlation with the number of vertices, increasing from 88.1\% for 3-vertex problems to 94.1\% for problems with 7 or more vertice. This trend validates that our synthesis process effectively controls for reasoning complexity.

\begin{table}[t]
\centering
\caption{Analysis of costs, failure rates (Qwen2.5-72B, CoT), and token lengths by vertex count for constructed Research Tree Data.}
\label{tab:vertex_analysis}
\renewcommand{\arraystretch}{1.2}
\sisetup{group-separator={,}, group-minimum-digits=4}
\setlength{\tabcolsep}{4pt} 
\begin{tabular}{l S[table-format=5.0] S[table-format=2.1] S[table-format=3.1] S[table-format=2.2] S[table-format=1.2]}
\toprule
\textbf{\# Vertices} & {\textbf{Count}} & {\textbf{Failure (\%)}} & {\textbf{Cost (\$)}} & {\textbf{Question Len (tok)}} & {\textbf{Answer Len (tok)}} \\
\midrule
3      & 3841 & 88.1 & 43.9 & 31.97 & 6.17 \\
4      & 15263 & 91.7 & 142.8 & 43.38 & 5.91 \\
5      & 15051 & 91.0 & 160.4 & 54.35 & 5.75 \\
6      & 17714  & 92.6 & 214.4 & 65.52 & 5.64 \\
$\geq$7   & 269   & 94.1 & 10.3   & 81.59 & 5.23 \\
\midrule
\textbf{Total} & \textbf{52138} & \textbf{91.6} & \textbf{571.8} & \textbf{53.43} & \textbf{5.79} \\
\bottomrule
\end{tabular}
\end{table}

\section{Method}
We introduce \textbf{InfoSeeker}, a framework for advanced agentic search and deep research, built upon our high-quality InfoSeek dataset. The core of our methodology is a novel workflow featuring parallel multi-query search and a dedicated Refiner Agent, designed to efficiently gather and synthesize vast amounts of information (Section~\ref{sec:workflow}). We then detail our two-stage training process: supervised fine-tuning (SFT) using rejection sampling to learn from successful reasoning trajectories (Section~\ref{sec:sft}), followed by reinforcement learning (RL) to further enhance its reasoning and search capabilities (Section~\ref{sec:rl}).

\subsection{Workflow with Multi-query Search and Refiner Agent}
\label{sec:workflow}
\begin{figure*}[t]
  \centering
  \vspace{-12pt}
  \adjustbox{center}{\includegraphics[width=1.05\textwidth]{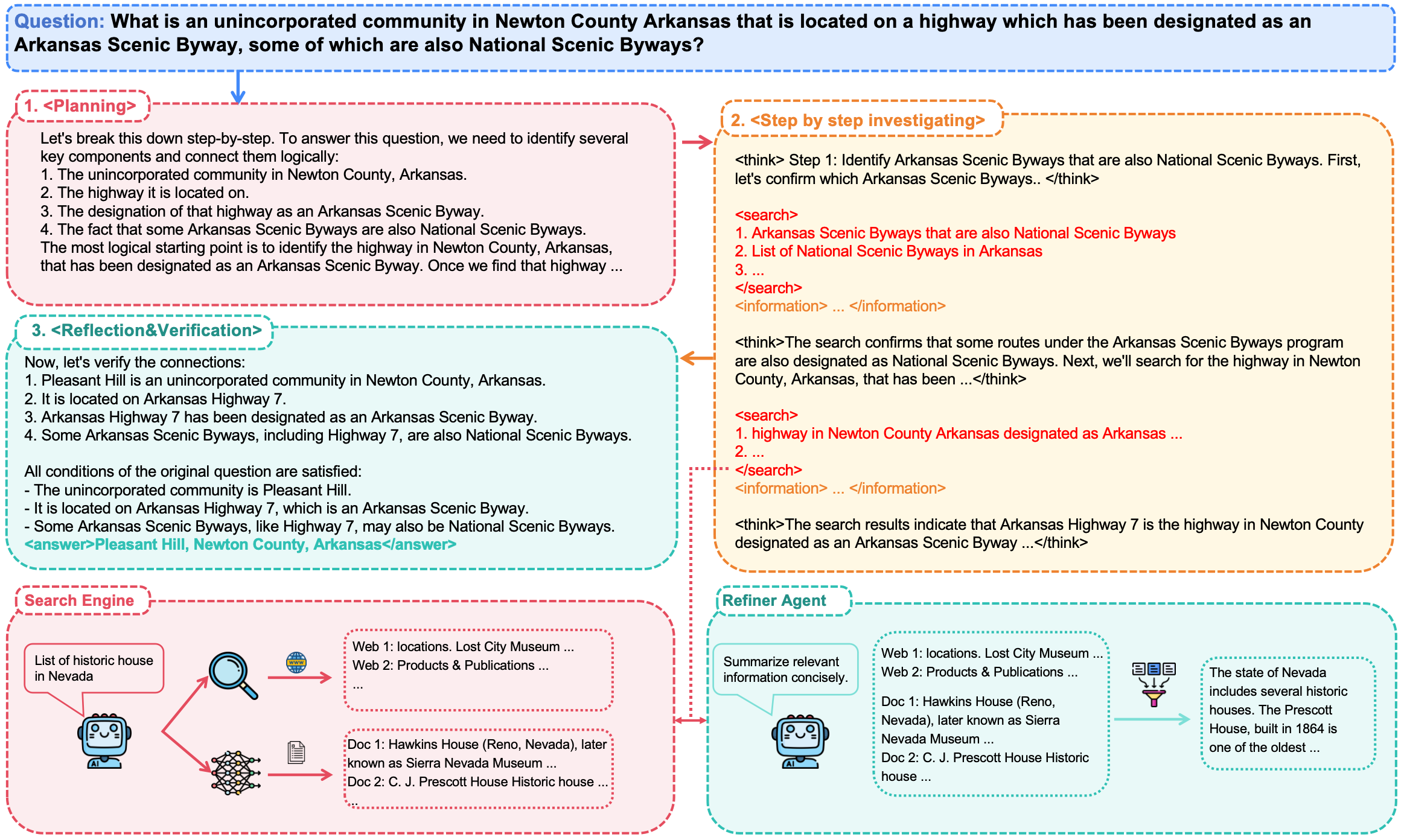}}
  \vspace{-4pt}
  \caption{Framework of InfoSeeker, which decomposes target tasks, conducts step-by-step investigation, and synthesizes final answer through coordination with the Search Engine and Refiner Agent.}
  \label{fig:workflow}
  \vspace{-10pt}
\end{figure*}

Recent studies~\citep{searchr1, chen2025learning, li2025websailor} have demonstrated the potential of LLMs to reason and invoke external search engines for complex tasks. A central challenge lies in effectively searching and integrating web information into the model’s context. Increasing snippet length or search depth (top-$k$) enhances recall by covering more information, but in multi-turn, agentic ReAct rollouts~\citep{yao2023react}, this quickly leads to bloated contexts filled with redundant or noisy evidence, causing the model to lose focus.
As shown in Fig.~\ref{fig:workflow}, we propose a novel workflow that addresses this limitation. At each step, InfoSeeker generates multiple queries in parallel, and a dedicated \emph{Refiner Agent} condenses the corresponding retrievals into concise summaries. This design maintains high recall while keeping the working context compact and tractable.

\textbf{Think Before Action.}  
Each reasoning turn in \textsc{InfoSeeker} begins with an explicit “thinking” phase, delimited by $\textcolor{ForestGreen}{\texttt{<think>}}$ and $\textcolor{ForestGreen}{\texttt{</think>}}$. This stage encourages the model to reflect on what has already been gathered and to plan what information remains necessary, leading to more targeted query generation and robust reasoning~\citep{yao2023react, jaech2024openai, guo2025deepseek}.

\textbf{Parallelized Multi-Query Search} 
To enhance the efficiency and broader coverage of information exploration, InfoSeeker generates multiple, diverse queries within a single reasoning step. These queries, enclosed within $\textcolor{Cyan}{\texttt{<search>}}$ and $\textcolor{Cyan}{\texttt{</search>}}$ tags, are designed to comprehensively address the current information-seeking intent from various angles. This parallelized approach broadens the informational coverage and accelerates the exploration process compared to a sequential, single-query strategy.

\textbf{Refiner Agent for Summarization.}
To enable InfoSeeker to efficiently exploit massive information from the multi-query search, we introduce a Refiner Agent.
For each query, the search engine returns the top-$k$ retrieved results, which are then passed to the Refiner Agent. The agent extracts salient evidence and produces a concise summary aligned with the query’s intent, while also offering recommendations for subsequent reasoning steps. Summaries are paired with their originating queries and encapsulated within $\textcolor{BlueViolet}{\texttt{<information>}}$ and $\textcolor{BlueViolet}{\texttt{</information>}}$.
In practice, we employ Qwen2.5-7B-Inst~\citep{qwen2} as the Refiner Agent, which is both efficient and effective enough across our validations.

\textbf{Output Answer.}  
Once the model determines that sufficient information has been accumulated, or the maximum number of search steps has been reached, it produces the final answer enclosed within $\textcolor{Dandelion}{\texttt{<answer>}}$ and $\textcolor{Dandelion}{\texttt{</answer>}}$. 

\subsection{Rejection Sampling as SFT}
\label{sec:sft}
A central challenge in developing deep research agents is navigating the vast exploration space of complex, multi-step reasoning tasks. Direct reinforcement learning is unstable and inefficient due to sparse rewards and a combinatorial action space, making it difficult for an agent to independently discover intricate workflows of planning, search, and verification in complex problems~\citep{tao2025webshaper, wu2025webdancerautonomousinformationseeking}.

To address this, we employ rejection sampling to construct a dataset of successful and executable reasoning trajectories for supervised fine-tuning (SFT). This approach filters for high-quality data by retaining only those trajectories that lead to correct outcomes. Specifically, we utilize both teacher model (Qwen2.5-72B~\citep{qwen2025qwen25technicalreport}) and preview version of InfoSeeker to solve tasks from our well-curated InfoSeek dataset with the workflow proposed in Sec~\ref{sec:workflow}. Only trajectories that successfully complete the task and yield a demonstrably correct final answer are accepted into our fine-tuning dataset. We further use Gemini 2.5 Flash~\citep{gemini25} to check if there exist search or reasoning shortcuts. SFT training details are described in Sec~\ref{sec:training_details} thoroughly.

\subsection{Reinforcement Learning with Reasoning and Search}
\label{sec:rl}
Reinforcement learning (RL) has been widely used in post-training phase to align models with human preferences~\citep{ouyang2022training} or strengthen LLMs' specific functionalities such as reasoning and tool using \citep{jaech2024openai, guo2025deepseek}. Previous works~\citep{searchr1, li2025websailor} have shown distinct effects on leveraging RL to enhance the capability in performaing complex searching tasks. We adopt an appropriate checkpoint from the SFT phase to start RL training, further make use of our constructed data to reinforce the model's ability in reasoning and writing precise queries.

\subsubsection{Algorithm}



\textbf{Group Relative Policy Optimization} \citep{shao2024deepseekmath} introduced Group Relative Policy Optimization (GRPO), a novel policy gradient based algorithm optimized for large scale training.

\vspace{-0.2cm}
\begin{align*}
    \mathcal{J}_{\text{GRPO}}(\theta)=\mathbb{E} \bigg[\frac{1}{G}\sum_{i=0}^G\frac{1}{|\mathcal{Y}|-|\mathcal{K}|}\sum_{t=1:I(\mathcal{Y}_t)=1}^{|\mathcal{Y}|}\min\left( r_{i,t} A_{i,t} \; , \; \text{clip}\left( r_{i,t}, 1-\epsilon, 1+\epsilon \right) A_{i,t} \right)\\
    -\beta\mathbb{D}_{KL}(\pi_\theta||\pi_{ref})\bigg]
\end{align*}
\vspace{-0.1cm}

As $r_{i,t}$ defined as $\frac{\pi_\theta(\mathcal{Y}_{i,t}|\mathcal{X},\mathcal{Y}_{i,<t},\mathcal{K}_{i,<t})}{\pi_{\theta_{\text{old}}}(\mathcal{Y}_{i,t}|\mathcal{X},\mathcal{Y}_{i,<t},\mathcal{K}_{i,<t})}$, GRPO adopts the clipped surrogate objective from PPO~\citep{schulman2017proximal}, while in company with a KL divergence penalty. Instead of using GAE~\citep{schulman2015high}, GRPO uses a new method to compute the advantages without value model. For each question input $\mathcal{X}$, we use the old policy model $\pi_{old}$ to generate a group of outputs $\{\mathcal{Y}_1, \mathcal{Y}_2, \dots, \mathcal{Y}_G\}$, and get $G$ rewards $\mathbf{R}=\{R_1, R_2, \dots, R_G\}$. Then the advantage is computed as a normalized reward: $A_{i,t}=\frac{R_i-\text{mean}(\mathbf{R})}{\text{std}(\mathbf{R})}$.

\subsubsection{Reward Design}

In reinforcement learning, reward serves as the fundamental signal that guides a model's learning and decision-making process. Instead of learning directly from labeled examples, the LLM generates responses and receives feedback in the form of rewards that indicate how desirable or useful those responses are. In our approach, since the model has already possess a degree of capability solving Deep Reserach tasks in desired format after the SFT phase, we design a relatively straightforward reward:
\[
R=
\begin{cases}
    1 &\text{if format and extracted answer are both correct}\\
    0 &\text{otherwise}
\end{cases}
\]
This binary design reward only those responses that both follow the required structure and contain the correct extracted answer, providing a clear signal to guide the model's optimization. RL training details are described in Sec~\ref{sec:training_details} thoroughly.

\begin{table*}[t]
\centering
\caption{Performance comparison on Single-Hop and Multi-Hop QA benchmarks. 
Best results in each column are highlighted in \textbf{bold}.}
\label{tab:results}
\renewcommand{\arraystretch}{1.20} 
\begin{tabularx}{1.02\linewidth}{lYYYYYYYY}
\toprule
\multirow{2}{*}{\textbf{Model}} & \multicolumn{3}{c}{\textbf{Single-Hop QA}} & \multicolumn{5}{c}{\textbf{Multi-Hop QA}} \\
\cmidrule(lr){2-4} \cmidrule(lr){5-9}
 & NQ & TQA & PopQA & HQA & 2Wiki & MSQ & Bamb & Avg. \\
\midrule
\rowcolor{gray!10} \multicolumn{9}{l}{\textbf{RAG-based Models}} \\
RAG & 34.8 & 54.4 & 38.7 & 25.5 & 22.6 & 4.7 & 8.0 & 27.0 \\
IRCoT & 11.1 & 31.2 & 20.0 & 16.4 & 17.1 & 6.7 & 8.0 & 15.8 \\
RQRAG & 32.6 & 52.5 & 39.4 & 28.5 & 30.7 & 10.1 & 12.9 & 29.5 \\
Self-RAG & 36.4 & 38.2 & 23.2 & 15.7 & 11.3 & 3.9 & 5.6 & 19.2 \\
\midrule
\rowcolor{gray!10} \multicolumn{9}{l}{\textbf{Agentic Search Models}} \\
Search-o1-3B & 23.8 & 48.2 & 26.2 & 22.1 & 21.8 & 5.4 & 32.0 & 25.6 \\
Searcn-R1-3B & 40.8 & 59.1 & 42.8 & 30.8 & 31.1 & 8.4 & 13.0 & 32.3 \\
ZeroSearch-3B & 41.2 & \textbf{61.5} & 44.0 & 31.2 & 33.2 & 12.6 & 14.3 & 34.0 \\
AutoRefine-3B & \textbf{43.6} & 59.7 & 44.7 & 40.4 & 38.0 & 16.9 & 33.6 & 39.6 \\
InForage-3B & 42.1 & 59.7 & 45.2 & 40.9 & 42.8 & 17.2 & 36.0 & 40.6 \\
\midrule
\rowcolor{gray!10} \multicolumn{9}{l}{\textbf{InfoSeeker}} \\
InfoSeeker-3B & 42.7 & 57.1 & \textbf{48.0} & \textbf{44.6} & \textbf{52.0} & \textbf{20.5} & \textbf{39.8} & \textbf{43.5} \\
\bottomrule
\end{tabularx}
\end{table*}

\begin{table}[t]
\centering
\caption{Model performance on the \textbf{BrowseComp-Plus} benchmark for complex reasoning tasks.}
\label{tab:browsecomp-plus}
\renewcommand{\arraystretch}{1.2}
{\setlength{\tabcolsep}{6pt}%
\begin{tabularx}{0.85\linewidth}{@{} >{\raggedright\arraybackslash}X c 
                                    S[table-format=2.1] 
                                    S[table-format=2.2] @{}}
\toprule
\textbf{Model} & \textbf{Retriever} & \textbf{Accuracy (\%)} & \textbf{Search Calls} \\
\midrule
Gemini 2.5 Flash   & BM25 & 15.5 & 10.56 \\
Gemini 2.5 Pro     & BM25 & 19.0 & 7.44 \\
Sonnet 4           & BM25 & 14.3 & 9.95 \\
GPT-4.1            & BM25 & 14.6 & 11.22 \\
GPT-5 & BM25 & 55.9 & 23.23 \\
Qwen3-32B          & BM25 & 3.5  & 0.92 \\
SearchR1-32B       & BM25 & 3.9  & 1.78 \\
\rowcolor{gray!12} InfoSeeker-3B  & BM25 & 16.5 & 8.24 \\
\bottomrule
\end{tabularx}}
\end{table}

\section{Experienments}
\subsection{Experiment Setting}
\paragraph{Datasets} We evaluate InfoSeeker on both single-hop benchmarks: Natural Questions (NQ) \citep{kwiatkowski2019natural}, TriviaQA (TQA)~\citep{TQA}, PopQA~\citep{PopQA}; and multi-hop benchmarks: HotpotQA (HQA)~\citep{HotpotQA}, 2WikiMultihopQA (2Wiki)~\citep{2WikiMultihop}, Musique (MSQ)~\citep{MuSiQue}, and Bamboogle (Bamb)~\citep{press2022measuring}. We use Exact Match (EM) is the evaluation metric for all these datasets. For evaluating more advanced deep research capability, we use the complex BrowseComp~\citep{browsecomp} benchmark, with the filtered 830 problems and fixed webpage corpus (100K) from BrowseComp-Plus~\citep{BrowseComp-Plus}. We use LLM to judge accuracy following the official setting.

\paragraph{Baselines} 
We compare with RAG-based methods: (1) Vanilla RAG, which retrieves top-k documents once and prepends them to the prompt; (2) IRCoT~\citep{IRCoT}, alternating retrieval with chain-ofthought reasoning; (3) RQRAG~\citep{chan2024rqraglearningrefinequeries}, which refines initial queries through rewriting and decomposition to improve retrieval accuracy; (4) Self-RAG~\citep{self-rag}, which introduces a self-reflection mechanism allowing the model to critique and revise its own outputs based on retrieved evidence.
We compare with more advanced agentic search methods: (1) Search-o1~\citep{searcho1}, which enhances LLMs with an agentic retrieval module and a Reason-in-Documents component for structured document reasoning; (2) Search-R1~\citep{searchr1}, which learns to generate multiple search queries during reasoning via reinforcement learning to optimize multi-turn retrieval interactions. (3) Zero-Search \citep{sun2025zerosearch}, which trains search agents using reinforcement learning without real search engines by simulating retrieval with another LLM. (4) AutoRefine \citep{shi2025search}, which introduces a "search-and-refine-during-think" paradigm using reinforcement learning with retrieval-specific rewards to iteratively filter and distill information before answering. (5) InForage \citep{qian2025scent} incorporate intermediate retrieval reward into agentic RL training.
For BrowseComp-Plus, we compare with the following models: Gemini 2.5 Flash, Gemini 2.5 Pro~\citep{comanici2025gemini}, Sonnet 4~\citep{anthropic2025claudeSonnet}, GPT-4.1~\citep{gpt-4}, GPT5~\citep{openai2025gpt5}, Qwen3-32B~\citep{yang2025qwen3}, Search-R1-32B~\citep{searchr1}.
The detailed evaluation details can be found in Appendix~\ref {sec:evaluation_details}.

\subsection{Main Results}
\paragraph{InfoSeeker exhibits strong general agentic search capability.} Table~\ref{tab:results} reports results on both single-hop and multi-hop QA benchmarks. We observe that InfoSeeker consistently outperforms all baselines, including both RAG methods and recent agentic search approaches. This indicates that our training pipeline enables the model to generalize effectively across diverse reasoning scenarios. Notably, most baselines rely heavily on large amounts of in-domain supervision (i.e., more than 100K NQ\&HQA), while our approach focuses on leveraging purpose-built InfoSeek dataset for training, yet still attains stronger general performance.

\paragraph{InfoSeeker exhibits strong deep research capability.} Table~\ref{tab:browsecomp-plus} highlights performance on the BrowseComp-Plus benchmark, which emphasizes open-ended, search-intensive reasoning. InfoSeeker-3B (16.5\%) surpasses several closed-source systems, including Gemini 2.5 Flash, Sonnet 4, and GPT-4.1, while also vastly outperforming open-source baselines such as Qwen3-32B and SearchR1-32B. Considering that InfoSeeker contains only 3B parameters, this result underscores the efficiency of our pipeline in scaling down deep research capabilities to compact LLMs.


\subsection{Comparison with Other Training Datasets}
We further evaluate the effectiveness of the InfoSeek dataset by comparing it with widely used datasets for agentic search, namely NQ and HotpotQA. Both datasets are frequently employed to train models with external information-seeking and multi-hop reasoning capabilities \citep{searchr1, sun2025zerosearch, shi2025search}. We conduct RL on NQ+HQA to compare with InfoSeek, using Qwen2.5-3B-Inst~\citep{qwen2025qwen25technicalreport} as backbone LLM. As shown in Table~\ref{tab:comparison}, training on InfoSeek yields substantially stronger deep research performance and more effective use of search tools.

\begin{table}[t]
\centering
\caption{Performance on BrowseComp-Plus with different training datasets.}
\label{tab:comparison}
\renewcommand{\arraystretch}{1.2}
{\setlength{\tabcolsep}{6pt}%
\begin{tabularx}{0.85\linewidth}{@{} >{\raggedright\arraybackslash}X c 
                                    S[table-format=2.1] 
                                    S[table-format=2.2] @{}}
\toprule
\textbf{Training Set} & \textbf{Retriever} & \textbf{Accuracy (\%)} & \textbf{Search Calls} \\
\midrule
NQ+HQA       & BM25 & 3.0  & 1.39 \\
InfoSeeker  & BM25 & 16.5 & 8.24 \\
\bottomrule
\end{tabularx}}
\end{table}

\section{Related Works}
\paragraph{Inference-Time Agentic Frameworks}
A significant body of work aims to enhance the problem-solving capabilities of pre-trained LLMs without altering their weights. These approaches construct agentic frameworks that operate at inference time, typically employing a central planning model to decompose complex problems and delegate sub-tasks to specialized tools or other LLM instances. For example, \citet{wu2025agentic} introduced Agentic Reasoning, a framework that dynamically assigns tasks to agents specializing in web search, coding, and memory management. Similarly, AgentOrchestra~\citep{zhang2025agentorchestra} proposes a hierarchical multi-agent system where a central planner delegates tasks to a suite of sub-agents. Other approaches, such as ALITA~\citep{qiu2025alita}, explore runtime self-evolution, enabling agents to dynamically generate and reuse tools on the fly. The primary focus of this research area is on the design of effective inference-time scaffolding and orchestration to leverage the existing capabilities of LLMs.

\paragraph{Training Agents for Search and Retrieval}
Another prominent research direction focuses on explicitly training agents to interact with external information sources, most notably search engines. Reinforcement learning (RL) is a common paradigm in this domain. For instance, R1-Searcher~\citep{song2025r1} and Search-R1~\citep{searchr1} both employ RL frameworks to teach LLMs how to interleave search queries with their reasoning steps, with the latter introducing techniques like retrieved token masking for more stable training. AutoRefine~\citep{shi2025search} presents a "search-and-refine" model where an agent learns to iteratively distill information from retrieved documents, guided by both retrieval-specific and final-answer rewards. Frameworks like Search-o1 \citep{searcho1} emphasize modularity by separating the search workflow from a document refinement module. The central goal of these works is to develop robust algorithms and policies for agents to effectively seek and utilize external knowledge.

\paragraph{Automated Data Synthesis for Deep Research Agents}
To cultivate more advanced reasoning skills, recent research has turned to automated data synthesis for training highly capable agents. These methods span a range of strategies. Some leverage RL within complex, open-ended environments; DeepResearcher~\citep{zheng2025deepresearcher}, for example, scales RL for agents interacting directly with the open web, while WebSailor~\citep{li2025websailor} synthesizes high-uncertainty web navigation tasks to train specialized agents. Other frameworks provide comprehensive ecosystems for agent development, such as Cognitive Kernel-Pro~\citep{fang2025cognitive}, or explore novel inference-time processes, like the iterative denoising approach of TTD-DR~\citep{han2025deep}.
A key challenge lies in ensuring the quality and logical consistency of the synthesized data. WebShaper~\citep{tao2025webshaper}, for instance, adopts a formalization-driven approach where a reasoning graph is defined before the corresponding question is generated. Our work contributes to this line of research on data synthesis, focusing on methods to automatically generate large-scale datasets with controllable and verifiable structural complexity to foster deep, multi-step reasoning.

\section{Conclusion}

We presented InfoSeek, a data-centric framework for advancing Deep Research with large language models. By formalizing verifiable Deep Research questions as \emph{Hierarchical Constraint Satisfaction Problems} (HCSPs), we established a principled distinction from simpler tasks such as multi-hop and flat CSP problems, and highlighted the need for high quality data that reflects hierarchical reasoning. InfoSeek operationalizes this formulation through a scalable data synthesis pipeline: entities and relations are mined from large-scale webpages, organized into Research Trees incrementally with sufficient constraints to blur parent vertices, ensuring that intermediate vertices form valid sub-problems and that solutions require traversing the full hierarchy. This design yields datasets that are structurally diverse, complexity-controllable, and intrinsically verifiable. Empirical evaluation confirms that models trained on InfoSeek dataset with supervised fine-tuning and reinforcement learning outperform strong baselines, validating its effectiveness for enabling more robust reasoning and tool use capabilities. Furthermore, because InfoSeek preserves meta-information such as intermediate steps and retrieval labels, it opens new opportunities for compound reward design and trajectory-level optimization. 


\bibliography{iclr2025_conference}
\bibliographystyle{iclr2025_conference}

\appendix
\section{Appendix}
\subsection{Training Details}
\label{sec:training_details}
In this section, we introduce the detailed training pipeline and implementation details of InfoSeeker. Training deep research agents is non-trivial, particularly for small scale LLMs. The key challenges lie in the scarcity of high-quality, complex data, and the lack of a clear, reproducible training pipeline. To address this, we construct the InfoSeek-50K dataset and design a two-stage training pipeline, enabling us to train a 3B LLM (Qwen2.5-3B-Inst)~\citep{qwen2025qwen25technicalreport} that approaches the performance of proprietary models.

\paragraph{Distill Teacher Model} Since small LLMs are inherently weaker, we begin with knowledge distillation from a larger teacher model. Specifically, we distill trajectories from Qwen2.5-72B~\citep{qwen2025qwen25technicalreport} executing research workflows proposed in Sec~\ref{sec:workflow}, which are then used for SFT of Qwen2.5-3B-Inst~\citep{qwen2025qwen25technicalreport}. Concretely, we utilize 50K InfoSeek samples (For training advanced deep research capability) and 5K NQ \& HQA samples (For preserving general agentic search capability), each rolled out twice. After filtering incorrect executions, we obtain 24K valid trajectories, implying that the teacher model achieves 21.8\% accuracy under our carefully designed workflow. Importantly, we deliberately retain “shortcut” cases among the correct trajectories, as preserving diverse solution strategies offers valuable learning signals for small LLMs during the early stages of training.

\paragraph{Two Round Training} Following distillation, we conduct Round 1 training to bootstrap the deep research capability of the backbone Qwen2.5-3B-Inst model. Using the 24K trajectories, we fine-tune the model for 2 epochs with a learning rate of 1e-5, weight decay of 0.01, and a context length of 16,384. Training on a single 8×H100 node completes in ~2 hours, yielding InfoSeeker-3B-SFT-Round1. We then perform reinforcement learning using GRPO with outcome-based rewards. Training is conducted with a batch size of 256, a maximum of 10 turns, rollout size 5, temperature 0.8, and a search engine restricted to the top-5 retrieved contents. After 200 RL steps, we obtain InfoSeeker-3B-RL-Round1.

To further strengthen deep research abilities, we proceed to Round 2 training. We first perform rejection sampling on InfoSeeker-3B-RL-Round1, generating 16,494 trajectories from 55K source samples. These are filtered via the Gemini 2.5 Flash~\citep{gemini25} API, producing 3,450 high-quality trajectories characterized by multi-turn search, finer-grained task decomposition, and more accurate step-by-step reasoning. The statistics are shown in Figure~\ref{fig:RFT}. Using the same hyperparameters as Round 1, we obtain InfoSeeker-3B-SFT. Next, we conduct the second stage of reinforcement learning. From the original 55K data pool, we select 17K harder samples (15K InfoSeek, 2K NQ \& HQA). Before training, the model generates preliminary answers, and we keep only 14K samples it fails on. Using GRPO \textbf{without KL loss} for 100 steps, we derive the final InfoSeeker-3B model.

\subsection{Evaluation Details}
\label{sec:evaluation_details}
For both single-hop and multi-hop QA tasks (NQ, TQA, PopQA, HQA, 2Wiki, MSQ, and Bamb), we employ Wikipedia-25 as the corpus, segmented into chunks of 512 tokens. Document retrieval is performed using BGE-M3~\citep{bge_m3}, with the top-$5$ documents selected.  
For the BrowseComp-Plus benchmark, we utilize the 100K web page corpus provided by the official release~\citep{BrowseComp-Plus}, with BM25~\citep{BM25} serving as the retrieval method.

\subsection{Case Study andFurther Statistics}
\label{sec:further_statistics}

\begin{figure*}[t]
  \centering
  \adjustbox{center}{\includegraphics[width=1.05\textwidth]{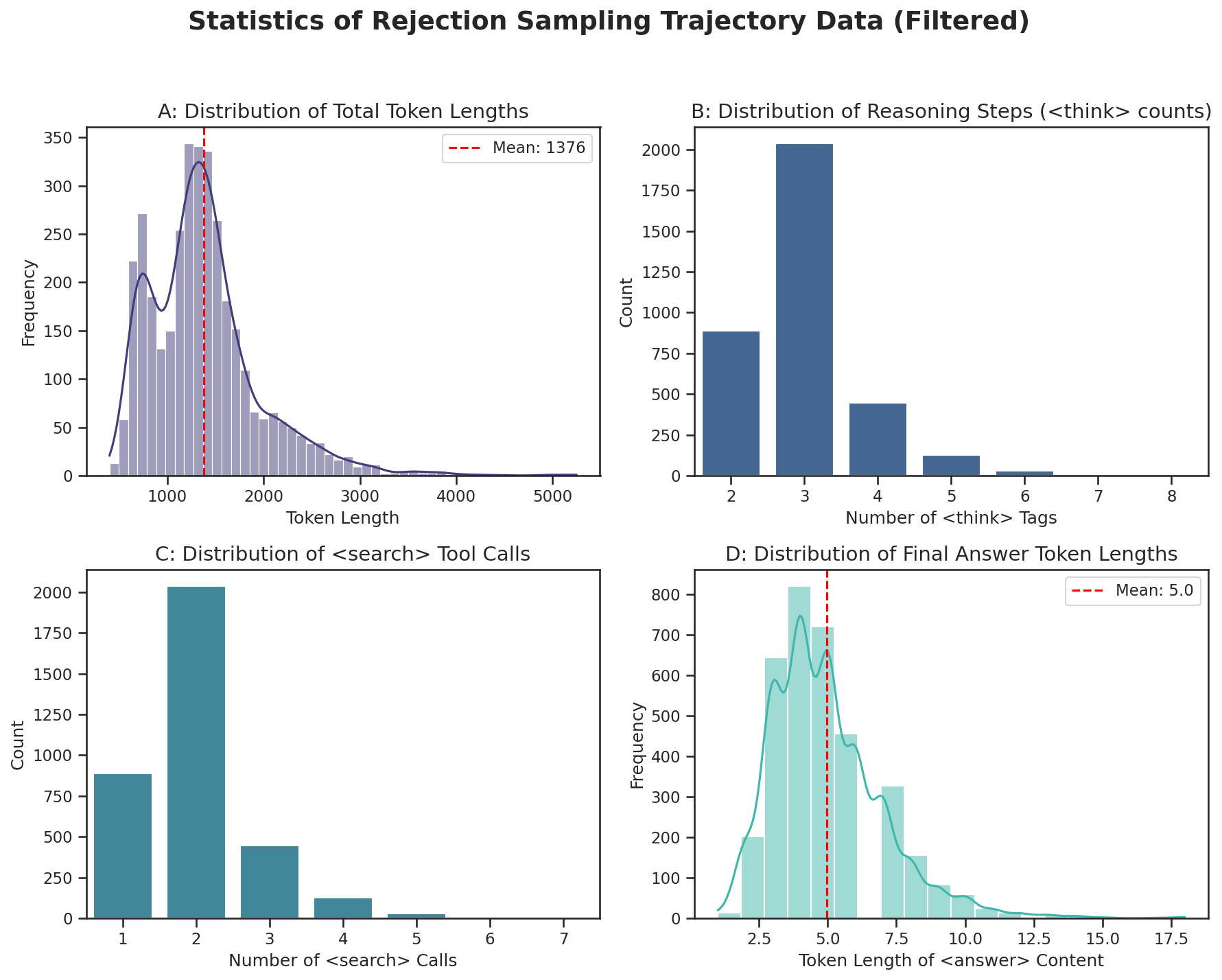}}
  \caption{Statistics for SFT trajectory data.}
  \label{fig:RFT}
\end{figure*}

\paragraph{SFT Trajectory Data} Figure~\ref{fig:RFT} provide detailed statistics for the constructed SFT Trajectory data from Research Tree data.

\begin{figure*}[t]
  \centering
  \adjustbox{center}{\includegraphics[width=1.05\textwidth]{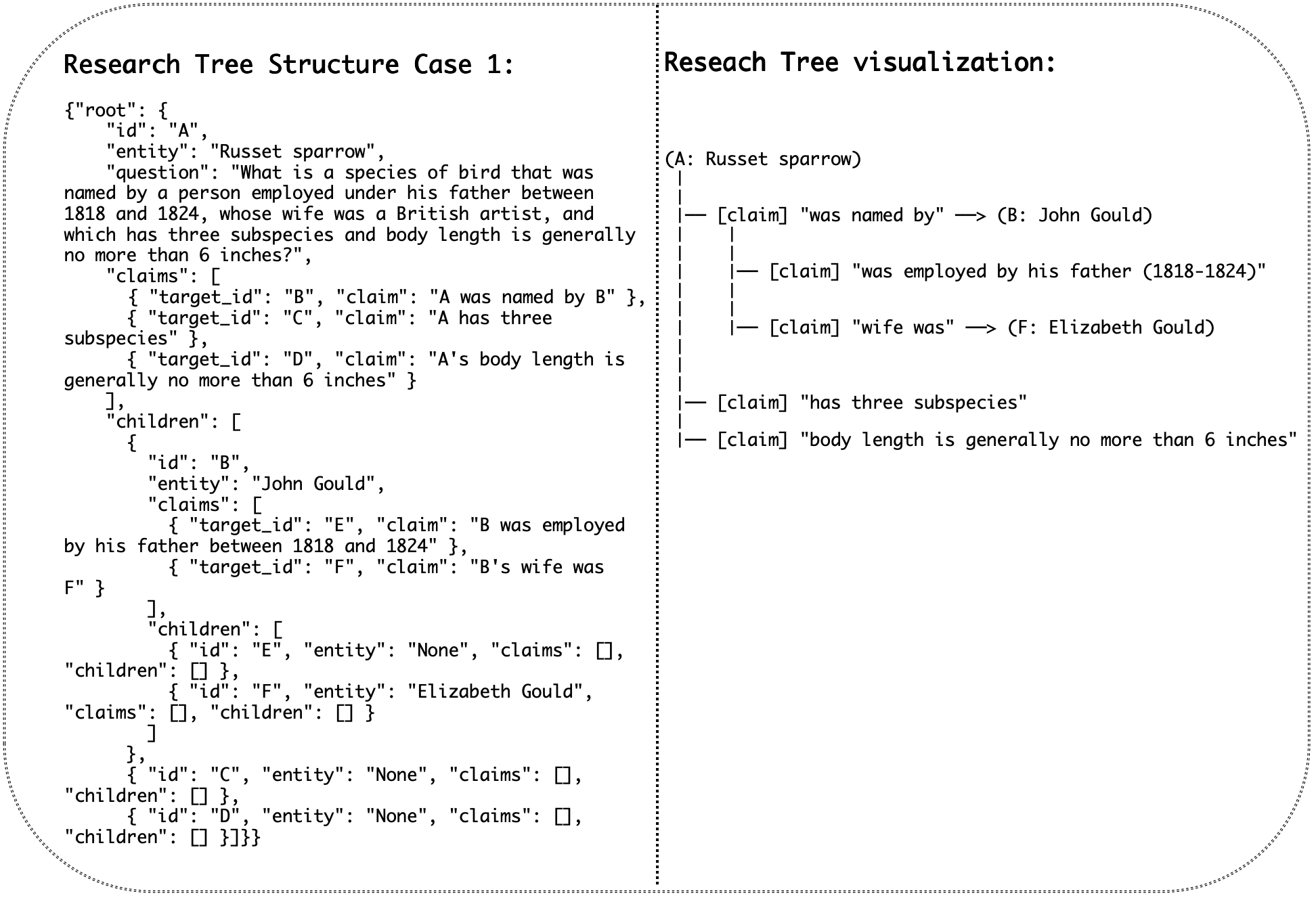}}
  \caption{Research tree structure of Case One in InfoSeek.}
  \label{fig:case_1}
\end{figure*}

\begin{figure*}[t]
  \centering
  \adjustbox{center}{\includegraphics[width=1.05\textwidth]{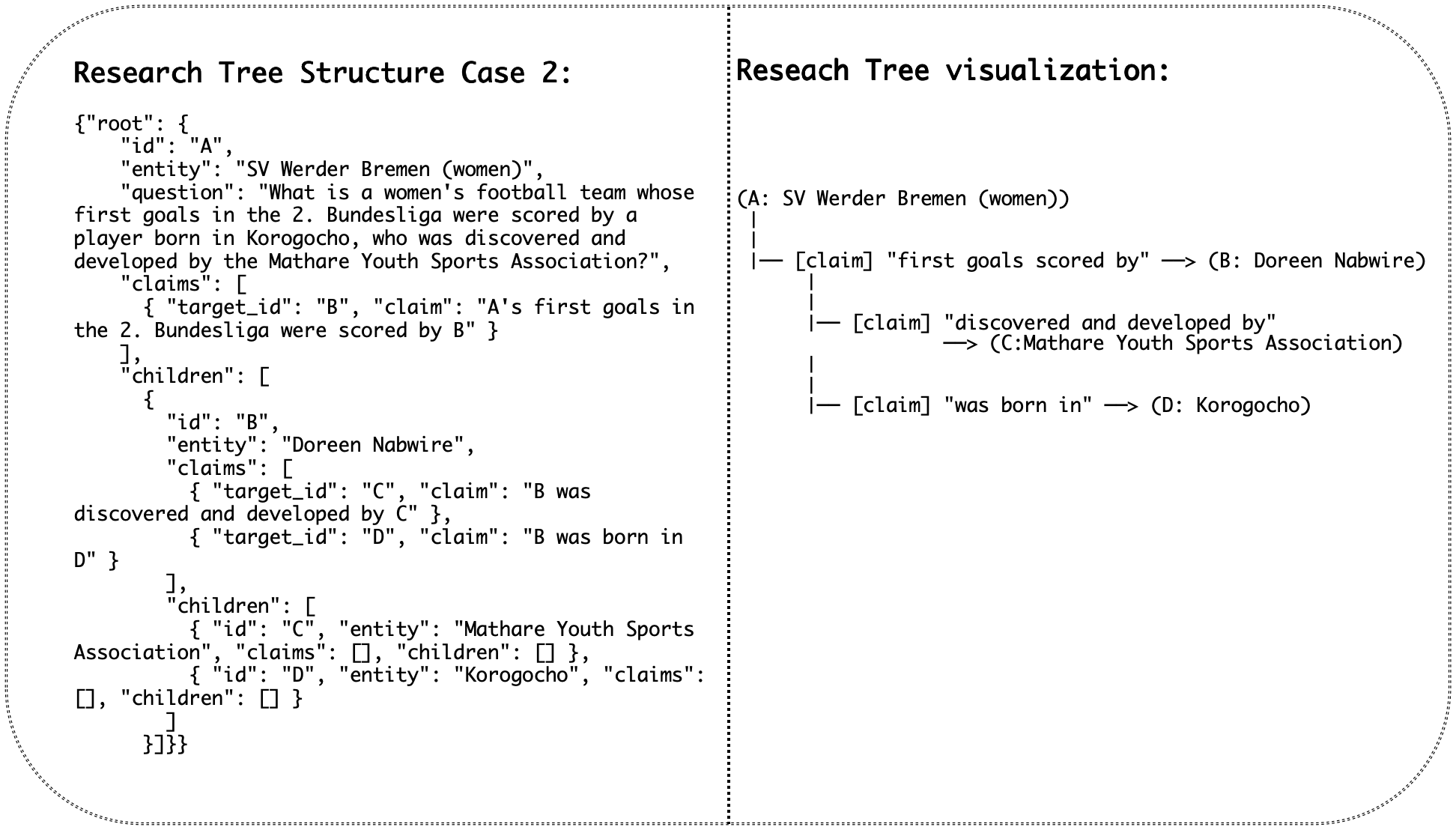}}
  \caption{Research tree structure of Case Two in InfoSeek.}
  \label{fig:case_2}
\end{figure*}

\paragraph{Case Study} Figure~\ref{fig:case_1} and Figure~\ref{fig:case_2} provide examples of constructed research tree structure and their visualization of InfoSeek.

\subsection{Announcement}
The code and data accompanying this work are released under the Apache License, Version 2.0. This permits use, modification, and distribution for research and commercial purposes, provided that proper attribution is given and the terms of the license are followed.

\end{document}